\documentclass[sigconf, nonacm]{acmart}

\def\BibTeX{{\rm B\kern-.05em{\sc i\kern-.025em b}\kern-.08emT\kern-.1667em\lower.7ex\hbox{E}\kern-.125emX}}

\begin{document}

	\title{What I See Is What You See: Joint Attention Learning for First and Third Person Video Co-analysis}
	
	%

	\author{
		Huangyue Yu$^{1}$,
		Minjie Cai$^{2}$,
		Yunfei Liu$^{1}$,
		Feng Lu$^{1,3,*}$
	}
	\affiliation{
	\institution{$^{1}$State Key Laboratory of VR Technology and Systems, School of Computer Science and Engineering, Beihang University, Beijing, China}
	\institution{$^{2}$Hunan University, Hunan, China}
	\institution{$^{3}$Peng Cheng Laboratory, Shenzhen, China}
}

	%
	%
	%
	%
	%
	
	%
	\renewcommand{\shortauthors}{Huangyue Yu, et al.}
	
	%
	\begin{abstract}
		In recent years, more and more videos are captured from the first-person viewpoint by wearable cameras. Such first-person video provides additional information besides the traditional third-person video, and thus has a wide range of applications. However, techniques for analyzing the first-person video can be fundamentally different from those for the third-person video, and it is even more difficult to explore the shared information from both viewpoints.
		In this paper, we propose a novel method for first- and third-person video co-analysis. At the core of our method is the notion of ``joint attention'', indicating the learnable representation that corresponds to the shared attention regions in different viewpoints and thus links the two viewpoints. To this end, we develop a multi-branch deep network with a triplet loss to extract the joint attention from the first- and third-person videos via self-supervised learning.
		We evaluate our method on the public dataset with cross-viewpoint video matching tasks. Our method outperforms the state-of-the-art both qualitatively and quantitatively. We also demonstrate how the learned joint attention can benefit various applications through a set of additional experiments.
		
	\end{abstract}

	%
	%

	%

	%

	\newcommand{\etal}{\emph{et al.}}

	\maketitle
	\section{Introduction}
	Nowadays, due to the widespread use of high-quality and low-cost digital cameras, a tremendous number of videos can be captured in every second. As a result, video analysis, which aims at automatically interpreting the visual content in the video, has become a research focus in the field of computer vision and multimedia processing. Its technical capability is useful or even crucial in a wide range of applications including object detection, action recognition, video surveillance and autopilot.
	
	Among all the videos captured in the real world, most of them are third-person videos. In other words, they are captured from a third-person viewpoint by the camera not associated with any person or object in the video. In contrast, first-person videos are usually captured by wearable cameras and see the visual world from a unique perspective which is inherently human-centric. Although the first-person videos are still a minority, its rapid increase in number has attracted attention from both industry and academia.

	\let\thefootnote\relax\footnotetext{*Corresponding Author: Feng Lu (lufeng@buaa.edu.cn).}

	\begin{figure}
		\begin{center}
			\includegraphics[width=1\linewidth]{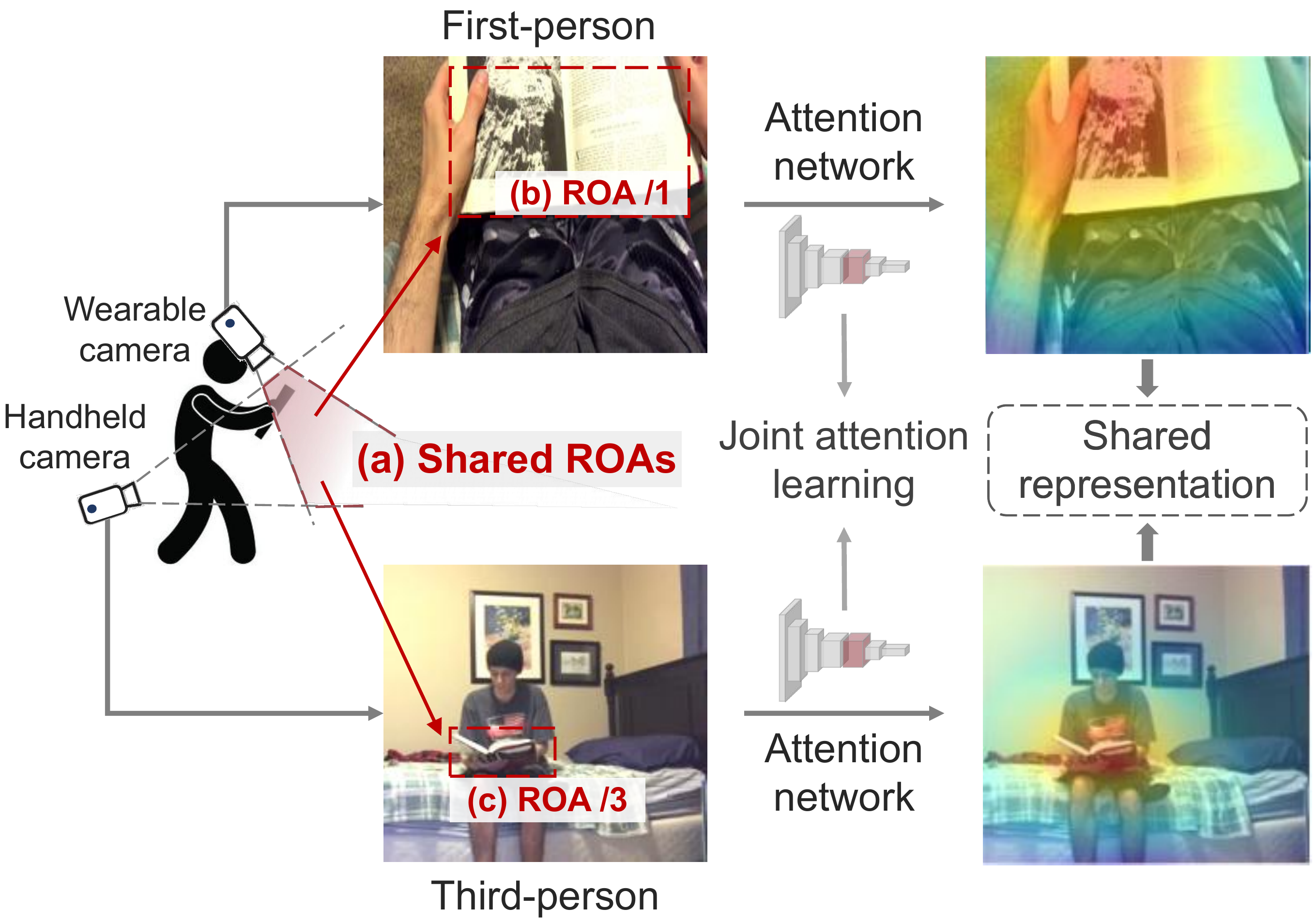}
		\end{center}
		\caption{Illustration of our motivation. (a) Shared regions of attention (shared ROAs, \emph{or joint attention regions}) between two viewpoints (highlight in red). (b) ROA in first-person video (highlight in red bounding box). (c) ROA in third-person video (highlight in red bounding box).}
		\label{teaser}
		\setlength{\abovecaptionskip}{-10cm}   
		\setlength{\belowcaptionskip}{-1cm}   
	\end{figure} 
	
	Both the first- and third-person videos are important since they capture the same real world from different perspectives. As a result, techniques have been proposed for analyzing either the first-person videos or the third-person videos respectively. However, it remains a challenge to explore information from both views in a joint manner. Very recently, Sigurdsson~\etal~\cite{sigurdsson2018actor} made an attempt to learn a shared representation of the two views by using a large dataset, which adopted the third-person information to help the first-person task. Although this work has demonstrated the benefit of linking the two views, its performance in finding the correct correspondences between them is still unsatisfactory. 
	
	The major difficulty lies in that the captured regions from the two views are quite different from each other --- the first-person view region only corresponds to a small and deformed part of that in the third-person view. Therefore, it is problematic to directly learn the shared representation of the two views' frames as in~\cite{sigurdsson2018actor}. Then the question has formed as how to find a more fundamental relationship between the two views to guide the shared representation learning, which enlightens our idea in conducting this research.

	In this paper, we propose to learn the shared representation from the first-person and third-person views in a more effective and robust way. Our key idea is to define and extract the ``joint attention'' between the two views, as shown in Fig.~\ref{teaser}. In particular, we define the region of attention (ROA) from both the first- and the third-person views, and make the assumption that the cross-view shared representation should correspond to the joint attention regions (shared ROAs). In this manner, our shared representation learning can further rely on the physically meaningful shared ROAs rather than the original video frames.
	
	In order to learn the joint attention as well as the shared representation, we propose a novel learning framework based on CNN. It incorporates the attention mechanism and self-supervised attention learning to extract shared ROAs from the first- and third-person videos without explicit annotations. First, attention maps of the two views' frames are generated respectively. Then, a self-supervised attention learning module is designed to extract the joint attention regions from both videos. Finally, our method transfers knowledge between the two views and applies the obtained shared representation to certain cross-view video co-analysis tasks. The method shall then benefit various related applications.

	Our main contributions include:
	\begin{itemize}
		\setlength{\itemsep}{2pt}
		\setlength{\parsep}{-2pt}
		\setlength{\parskip}{0pt}
		
		\item We introduce the joint attention, based on which the shared representation between the first- and third-person videos can be learned more effectively and robustly.
		
		\item We propose a new multi-branch deep network for joint attention learning and shared representation learning, based on a self-supervised attention learning architecture. The network extracts important features from cross-view videos effectively.
		
		\item Our method outperforms the state-of-the-art in cross-view video matching tasks. Additional experiments are also presented to show how our method is beneficial to various related applications.

	\end{itemize}

	\section{Related Works}
	
	\textbf{Modeling between first- and third-person videos.}
	Recent studies of modeling between first- and third-person videos have been often conducted on paired videos of these two domains~\cite{yonetani2015ego,ardeshir2016ego2top,fan2017identifying,sigurdsson2018actor,xu2018joint}.
	Yonetani~\etal ~\cite{yonetani2015ego} proposed a novel face detection approach by matching camera and head motion of the same person from first- and third-person perspective.
	Ardeshir and Borji
	~\etal ~\cite{ardeshir2016ego2top} matched a set of first-person videos to a set of characters (first-person camera wearers) in a top-view surveillance video using graph matching.
	Fan~\etal ~\cite{fan2017identifying} studied a similar problem by learning a joint feature embedding space from first- and third-person videos with a two-stream semi-Siamese network.
	Unlike~\cite{fan2017identifying} which requires ground-truth human bounding boxes, Xu~\etal ~\cite{xu2018joint} simultaneously segmented and matched the first-person camera wearers in third-person videos.
	
	While the above works assumed that the paired first- and third-person videos are synchronized, we propose an approach to temporally match individual video frames between these two domains. The most related work to ours is that of Sigurdsson~\etal ~\cite{sigurdsson2018actor}, which learned a joint embedding space between matched first- and third-person video frames. Different from ~\cite{sigurdsson2018actor}, we propose to learn joint attention in both videos for more accurate matching.
	
	\textbf{Attention model.}
	Detecting and understanding the attention regions in images and videos have been an emerging research field in computer vision and multimedia processing these years. Attention model has shown its efficiency in various vision tasks such as person re-identification ~\cite{Lejbolle2018Attention,Li2018Diversity,Li2018Harmonious,Song2018Mask}, image captioning ~\cite{Chen2018Factual,Chen2018Boosted,Anderson2018Bottom-Up}, pose estimation ~\cite{Wentao2018Cascaded,Parisotto2018Global,Chu2017Multi} and image classification ~\cite{Y2018Object,Y2018Multi}. In 1998, Itti~\etal ~\cite{itti1998model} constructed a primary visual attention model using a bottom-up architecture. After that, various attention models with different architectures are inspired: Wang~\etal~\cite{Wang2018Non-Local} proposed a non-local blocks operation,  which is related to the self-attention method, and computes the response at a position as a weighted sum of the features at all positions. 
	In~\cite{Du2018Interaction}, an interaction-aware attention network was presented to construct a spatial feature pyramid for obtain more accurate attention maps by multi-scale information. 
	Woo~\etal~\cite{Woo2018CBAM} made an extension of SE module~\cite{Hu2018Squeeze}, and presented a CBAM attention model. Jun~\etal ~\cite{Jun2019Dual} introduced visual attention into image segmentation, which then achieved better performance through their long-range context relationships.
	
	Different from the previous works, we use data from different viewpoints to jointly learn the shared ROAs from videos. The shared ROAs will more focus on the joint attention regions from different viewpoints.
	
	\textbf{Cross-view representation learning.} 
	Existing research on cross-view representation learning usually adopt deep metric learning with siamese (triplet) architectures~\cite{Hu2018CVM,Nam2016Localizing} or propose an encoder-decoder framework with generative adversarial networks~\cite{Regmi2018Cross,Bo2018Multi}. Regmi~\etal ~\cite{Regmi2018Cross} addressed the novel problems of cross-view image synthesis, aerial to street-view and vice versa, by using conditional generative adversarial networks to learn shared representation. Hu~\etal ~\cite{Hu2018CVM} used a triplet loss component, and proposed a CVM-Net for ground-to-aerial geolocalization. 
	
	In this paper, the notion of ``joint attention" is to be developed, which holds the view that the shared representation among cross-viewpoints should correspond to the joint attention regions. We construct a self-supervised attention learning architecture to extract the joint attention region from cross-viewpoints.

	\section{Joint attention guided representation learning}
	
	
	\subsection{Study of Attention Mechanism in Shared Representation Learning}
	The goal of this paper is to learn a shared representation for co-analysis of first- and third-person videos. 
	However, the learning of shared representation for videos from different viewpoints is challenging, especially for the first- and third-person videos. Due to fundamentally different viewpoints in data recording, the data from first-person videos varies a lot from that in third-person video. Assuming a pair of first- and third-person videos recording the same action, while a first-person video records detailed hand-object interactions from an actor's own viewpoint, a third-person video records the actor's whole body pose and surrounding environment from an observer's viewpoint. The different between these two makes straightforward learning of shared representation across the two viewpoints difficult.
	
\begin{figure}
		\begin{center}
			\includegraphics[width=1\linewidth]{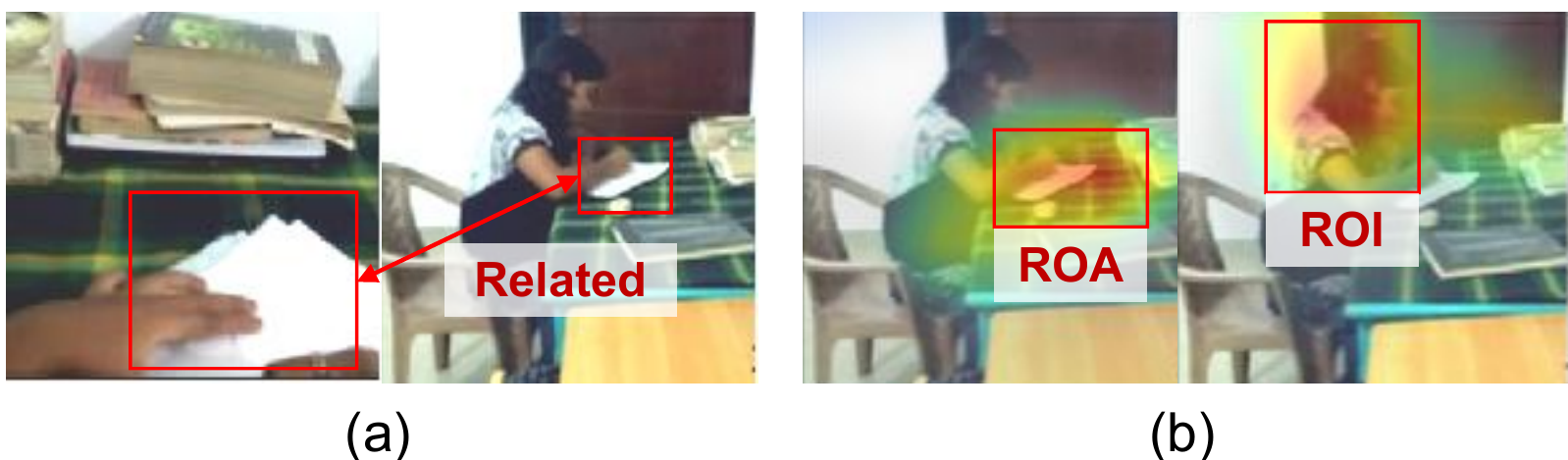}
		\end{center}
		\caption{The motivation of self-supervised attention learning. (a) shows a pair of first- and third-person frames, and the red bounding box indicates the shared ROAs from two viewpoints. (b) displays the difference between region of attention (ROA) and region of interest (ROI).}
		\label{AttentionSaliency}
		\vspace{-0.5cm}  
		\setlength{\abovecaptionskip}{-1cm}   
		\setlength{\belowcaptionskip}{-0.5cm}   
	\end{figure}

	As studied in previous works~\cite{Wang2017Residual,Hu2018Squeeze}, attention in an image tells the computer the focus regions and then to suppress unnecessary information. As shown in Figure~\ref{AttentionSaliency} (a), although the two images contain different contents, the regions within red bounding boxes of both images share consistent appearance and could be exploited to extract a shared representation for both images. Here we use region of attention (ROA) to denote the informative region about the recorded action. In order to reliably learn the shared representation between first- and third-person videos from quite different viewpoints, it is important to consider ROAs shared between the two videos. Therefore, we propose to adopt attention mechanism to estimate ROAs from the first- and third-person videos based on the shared representation learned.
	
	
	\subsection{Self-supervised Joint Attention Learning}
	Although the incorporation of attention mechanism could help the learning of shared representation between first- and third-person videos, identifying the regions of attention is a nontrivial task. Traditional saliency-based models tend to locate the region of interest (ROI) that is attractive to human perception~\cite{Zhang2018deep} but not necessarily correspond to the informative region (or ROA) desired for shared representation learning.
	
	Instead of directly applying the extracted ROI from previous works, we aim to learn ROA which corresponds to the informative region of first- and third-person videos and therefore is more suitable for shared representation learning. Motivated by the observation that the shared ROAs (or \textit{joint attention} regions) between first- and third-person videos contain consistent appearance, we propose a self-supervised learning approach to learn the joint attention regions by guaranteeing the semantic consistency between representations of ROAs of the first- and third-person videos. We visualize the ROA generated by our method and the ROI of a third-person video frame in Fig.~\ref{AttentionSaliency} (b). While ROI is more likely to focus on the visually interesting regions of human body, our learned ROA concentrates more on the person's action field and contains more consistent information with the first-person viewpoint.

	\subsection{Simultaneous Learning of Joint Attention and Shared Representation}
	
	As studied in previous sections, the extraction of joint attention regions benefits the learning of shared representation between first- and third-person videos. Meanwhile, shared representation helps determine whether two video frames of different viewpoints are corresponding to each other and thus provides a constraint of semantic consistency for self-supervised learning of joint attention. Therefore, the learning of joint attention and shared representation are interdependent and should be solved together.
	
	Given a triplet of frames as input, which contains a third-person video frame, a corresponding first-person video frame, and a non-corresponding first-person video frame, our learning task is to extract ROAs and feature representations for each frame which satisfy the following two constraints: (1) ROAs are of semantic consistency between corresponding frames, and (2) feature representations are close between corresponding frames and distant between non-corresponding frames. 
	
	With an objective function considering both constraints above, we train a multi-branch convolutional neural network to simultaneously learn joint attention and shared representation between first- and third-person videos. The technical details are given in the next section.

	\begin{figure}
		\begin{center}
			\includegraphics[width=\linewidth]{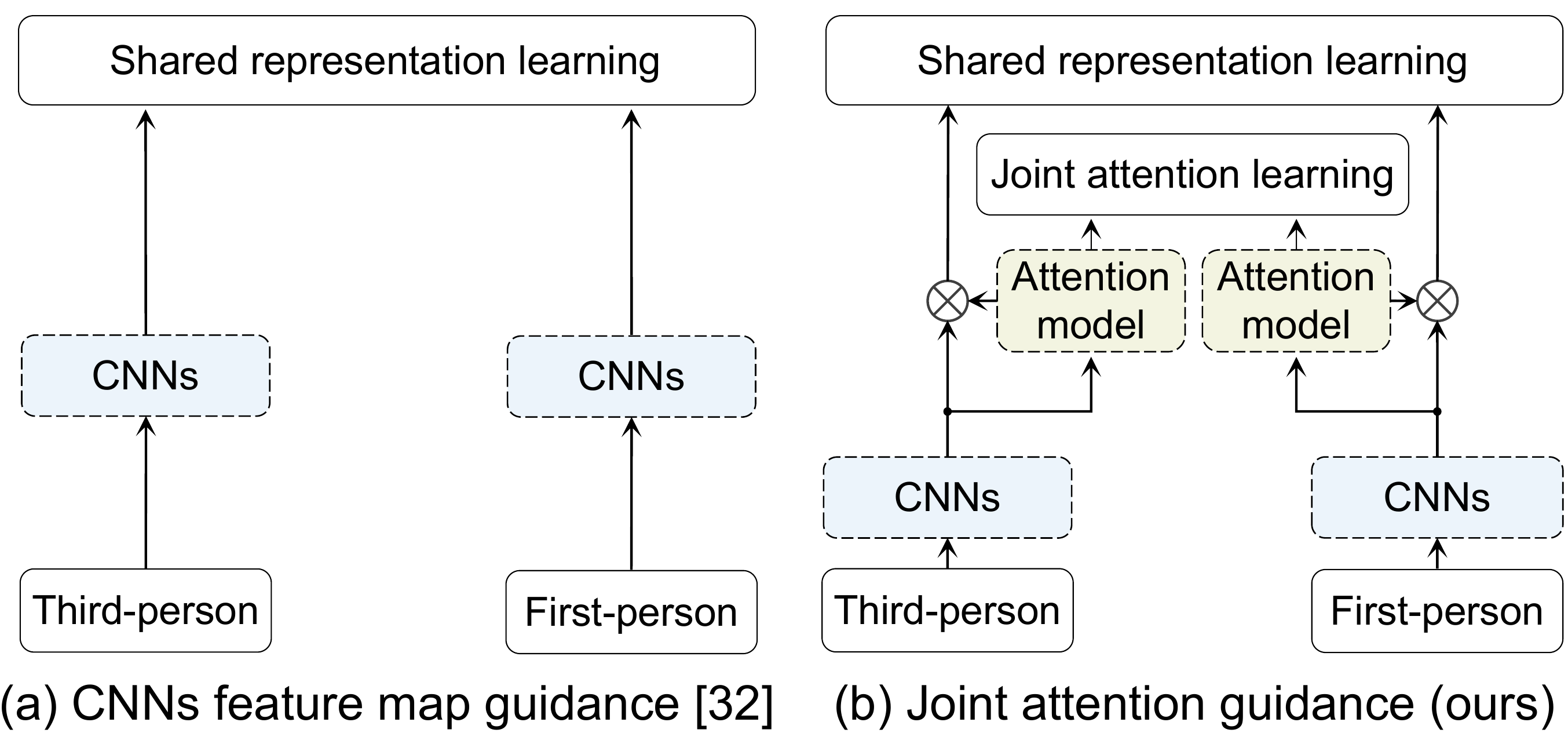}
		\end{center}
		\caption{Pipeline comparison between AONet~\cite{sigurdsson2018actor} and ours. (a) AONet~\cite{sigurdsson2018actor} learns the shared representation based on CNNs feature map directly. (b) Our architecture adopts joint attention learning module as guidance to filter the CNN-based features for learning of shared representation.}
		\label{OurNetwork2} 
		
		\vspace{-0.5cm}  
		\setlength{\abovecaptionskip}{-1cm}   
		\setlength{\belowcaptionskip}{-0.5cm}   
	\end{figure}

	\begin{figure*}[t]
		\begin{center}
			\includegraphics[width=1\linewidth]{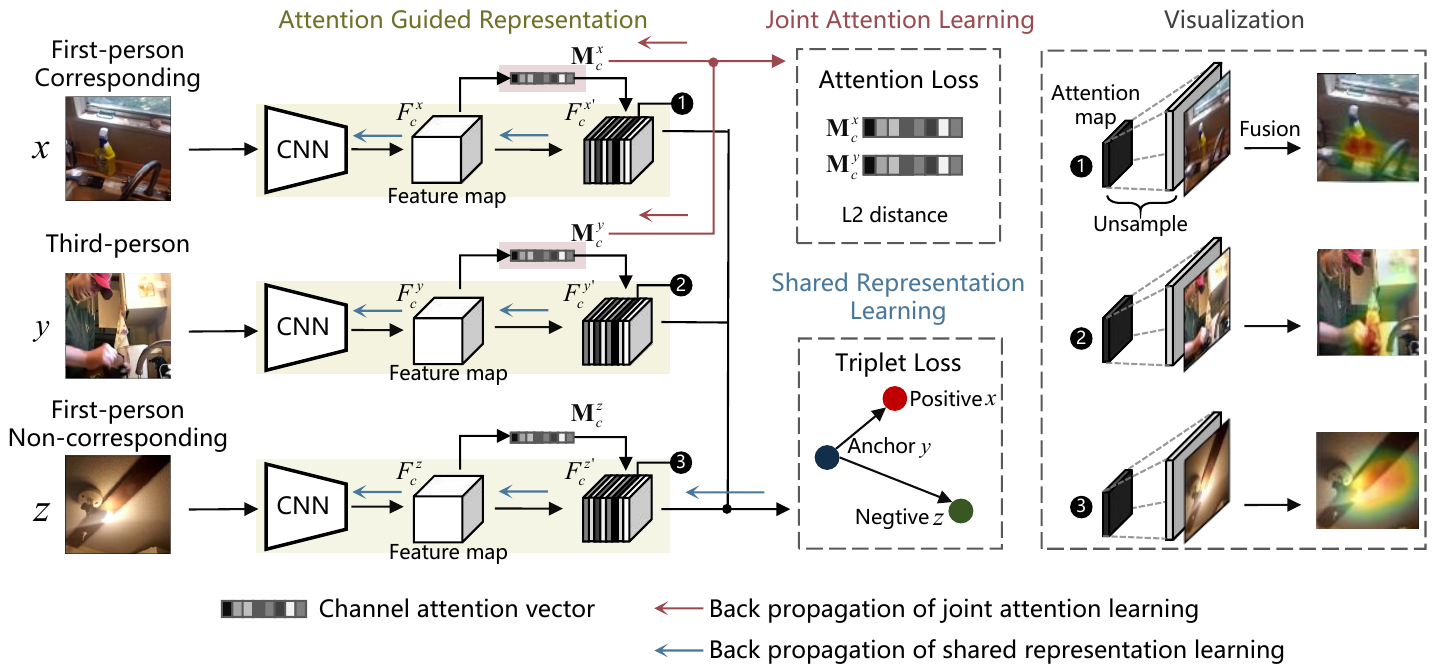}
		\end{center}
		\caption{The architecture of our proposed network. Taking a triplet frames $(x,y,z)$ as input, an attention-guided feature extraction module is developed to extract channel attention vectors $(\mathbf M_{c}^{x},\mathbf M_{c}^{y},\mathbf M_{c}^{z})$ from the intermediate feature maps $({F_{c}^{x}},{F_{c}^{y}},{F_{c}^{z}})$. Then, the self-supervised joint attention learning module encourages similarity the attention vectors between third-person frame $\mathbf M_{c}^{y}$ and corresponding first-person frame $\mathbf M_{c}^{x}$ to predict joint attention regions (shared ROAs), with the assumption that shared representation should correspond to the joint attention regions. Finally, the shared representation learning module explores the information between two viewpoints to obtain the shared representation. We weight average attention maps based on the channel to visualize what regions the model are learning from, the results are shown on the right.}
		\label{OurNetwork}
	\end{figure*}	
	
	\section{Methodology}
	
	\subsection{Overview}
	In this paper, we aim to learn a shared representation for co-analyzing the first- and third-person videos. In contrast with previous works that learned shared representation straightforwardly from videos, we incorporate attention mechanism into the representation learning framework. In particular, we propose a self-supervised joint attention learning module to predict joint attention regions for shared representation learning based on the assumption that shared representation should correspond to the joint attention regions from videos of different viewpoints. 
	
	The pipeline of our method is demonstrated in Fig.~\ref{OurNetwork2} (b).
	At first, basic features are generated from video frames of different viewpoints by a standard CNN. Then, instead of directly comparing the CNN-based features from two viewpoints, a novel joint attention learning module is proposed to predict ROA for each viewpoint. Finally, the predicted ROAs are used as guidance to filter the CNN-based features for learning of shared representation. 
	
	\subsection{Architecture}
	The architecture of our proposed framework is presented in Fig.~\ref{OurNetwork}. The framework is composed by a multi-branch neural network and takes a triplet of frames $(x,y,z)$ as input. It contains three main modules: attention-guided feature extraction, self-supervised joint attention learning and shared representation learning, which are described in the following parts.
	
	\textbf{Attention-guided feature extraction}. The purpose of this module is to focus on informative regions and suppress redundant features for shared representation learning at the later stage. A base CNN model is used to extract feature maps $({F_{c}^{x}},{F_{c}^{y}},{F_{c}^{z}})$ for each input branch. The parameters of the base CNN model are shared by all branches. A state-of-the-art attention model~\cite{Woo2018CBAM} is used to generate channel attention vectors which model relative importance of different semantic information represented by the feature maps. The channel attention vectors are denoted as $(\mathbf M_{c}^{x},\mathbf M_{c}^{y},\mathbf M_{c}^{z})$, where $\mathbf M_{c}\in \mathbb{R}^{c\times 1\times 1}$ and $c$ indicates channel number. 
	The original feature maps $({F_{c}^{x}},{F_{c}^{y}},{F_{c}^{z}})$ are then filtered by the generated channel attention vectors and the final feature representations $({F_{c}^{x}}',{F_{c}^{y}}',{F_{c}^{z}}')$ are defined as:
	\begin{equation}
	{F_{c}}' = \mathbf M_{c}(F_{c})\otimes F_{c},
	\end{equation}
	where $\otimes$ denotes element-wise multiplication.

	\textbf{Self-supervised joint attention learning}. In this module, we aim to learn joint attention regions between first- and third-person videos which focus on the same action recorded from two viewpoints. We develop a self-supervised learning approach by enforcing the semantic consistency between ROAs of corresponding video frames. The inputs are the pairs of channel attention vectors $(\mathbf M_{c}^{x},\mathbf M_{c}^{y})$ generated from a third-person video frame and the corresponding first-person video frame. We compare the two vectors with a L2-based distance metric, which enforces similarity between channel attention vectors of corresponding frames and thus guarantees the semantic consistency of ROAs between first- and third-person videos.
	
	\textbf{Shared representation learning}. In this module, we learn shared representation between first- and third-person videos based on the feature representations $({F_{c}^{x}}',{F_{c}^{y}}',{F_{c}^{z}}')$ that are filtered by the predicted attention at the previous stage. We adopt a triplet loss to enforce that the feature representations of corresponding frames ($x$ and $y$) are close to each other, and verse otherwise.
	
	\subsection{Loss Function}
	Here we describe the objective function used to train our proposed network. The objective function consists of two loss functions: an attention loss used to learn joint attention and a triplet loss used to learn shared representation.
	
	The attention loss is denoted as  $\mathcal{L_{\mathit{AL}}}(x,y)$, which is based on L2 distance metric to enforce similarity between channel attention vectors of the corresponding first- and third-person video frame. It is formulated as: 
	\begin{equation}
	\mathcal{L_{\mathit{AL}}}(x,y)= \left \| \mathbf M_{c}^{x} - \mathbf M_{c}^{y} \right \|_{2}.
	\end{equation}
	where $\left \| \cdot  \right \|_{2}$ denotes the L2 distance metric.
	
	The triplet loss is denoted as $\mathcal{L_{\mathit{TL}}}(x,y,z)$, which enforces similarity between corresponding feature representations ${F_{c}^{x}}'$ and  ${F_{c}^{y}}'$, and penalizes similarity between non-corresponding feature representations ${F_{c}^{y}}'$ and ${F_{c}^{z}}'$. It is formulated as:
	\begin{equation}
	\mathcal{L_{\mathit{TL}}}(x,y,z) = \frac{e^{{\left \| {F_{c}^{x}}' - {F_{c}^{y}}' \right \|}_{2}}   }{e^{{\left \| {F_{c}^{x}}' - {F_{c}^{y}}' \right \|}_{2}}   -e^{{\left \| {F_{c}^{y}}' - {F_{c}^{z}}' \right \|}_{2}}},
	\end{equation}
	
	Following~\cite{sigurdsson2018actor}, we also assign importance weight $w(x,y,z)$ to each triple to remove the negative impact of uninformative frames that are recorded under unstable conditions.
	
	The final loss  $\mathcal L(x,y,z)$ is organized: 
	\begin{equation}
	\mathcal L(x,y,z)=[\mathcal{L_{\mathit{TL}}}(x,y,z)+\lambda \cdot 	\mathcal{L_{\mathit{AL}}}(x,y)]\cdot  w(x,y,z),
	\end{equation}
	where $\lambda$ is a hyper parameter used to balance the relative contributions of different losses. We empirically set $\lambda=2.5$ in our experiments.
	
	\subsection{Implementation Details}
	Our framework is implemented by using PyTorch~\cite{paszke2017automatic}, and input frames are crop into $224 \times 224$. We apply a ResNet-152 architecture~\cite{Kaiming2016Deep} as the base CNN model, which is pretrained on Charades dataset~\cite{Gunnar2016Hollywood}. The first four convolutional layers of ResNet-152 are used to extract feature maps. We implement our channel attention module following the same settings of CBAM~\cite{Woo2018CBAM}. To reduce parameter overhead, the hidden activation size is $\mathbb{R}^{c/r\times 1\times 1} $ and the reduction radio $r$ is set as 8. The channels of feature maps and the dimension of channel attention vectors are both 2048. The spatial size of feature maps as well as that of attention maps is $7 \times 7$. The triplet importance weight is learned from one fully connected layer based on the fc7 features of ResNet-152. SGD is used to train the whole model, with the learning rate of 3e-5 and batch size of 4.

	\section{Experiment}
	In this section, we evaluate our method on a public dataset with pairs of first- and third-person videos. We first conduct performance comparison with state-of-the-art method and ablation study to verify the effectiveness of our method both qualitatively and quantitatively. We also conduct additional experiments to demonstrate the benefits of our work on various applications. 
	
	\begin{figure}
		\setlength{\belowcaptionskip}{-0.3cm}   
		\begin{center}
			\includegraphics[width=1.0\linewidth]{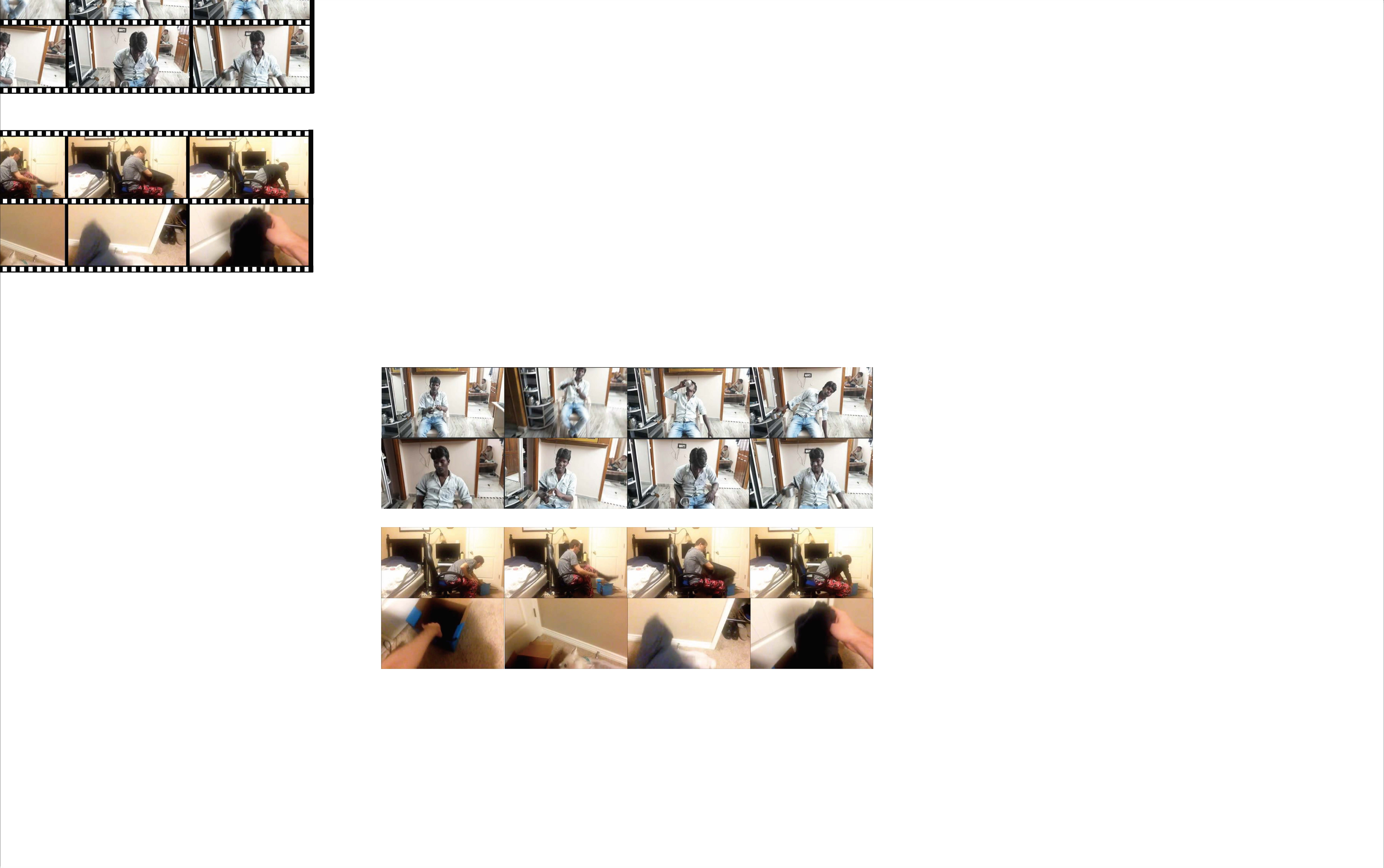}
		\end{center}
		\caption{Examples of invalid video pairs (top) and valid video pairs (bottom).}
		\label{IncorrectPerspective}
	\end{figure}
	
	\subsection{Experimental Settings}
	\textbf{Dataset}. To evaluate our method, we introduce Charades-Ego dataset which involves 4000 paired first- and third-person videos of daily indoor activities recorded by 112 persons.  
	Since our goal is to learn a shared representation between first- and third-person videos, we focus on the video pairs that record the same action from the first- and third-person viewpoints respectively. However, there are some video pairs in the raw dataset of which the viewpoints are invalid for our research. As shown in Fig.~\ref{IncorrectPerspective}, while the bottom half part illustrates image samples of a valid video pair, the top half part illustrates an invalid video pair in which an action is recorded from slightly different third-person viewpoint. Therefore, we carefully examined the dataset and removed 189 invalid video pairs in total.
	
	\textbf{Tasks and evaluation metrics}. We compare our method with different baselines on two cross-view video matching tasks: pairs discrimination and best-match moment localization. 
	
	The task of pairs discrimination aims to discriminate corresponding first- and third-person image pairs from the non-corresponding ones. Giving a triplet of third-person frame with corresponding and non-corresponding first-person frames, we identify one of the two first-person frames as the corresponding frame which has closer distance of representation with the third-person frame. In this task, we use classification accuracy as the evaluation metric. 
	
	The task of best-match moment localization aims to find the corresponding moment (1 second video clip) in a video given a video moment from the other viewpoint without the knowledge of time stamp. We use average distance of feature representation between moments of two viewpoints to identify the best match moment. For evaluation, we assume that the ground truth best-match moment can be approximated by temporally scaling the first-person video to have the same duration as the third-person video.
	Temporal alignment error is used as the evaluation metric for this task.

	\textbf{Comparison with state-of-the-art}. We compare our method with the state-of-the-art method ActorObserverNet (AONet)~\cite{sigurdsson2018actor}. AONet learns a shared representation between first- and third-person videos with a Siamese-like network. We re-train their network with default parameters based on our filtered dataset.
	
	\textbf{Ablation study}. To evaluate how different parts of our model contribute to the final performance on the two tasks, we conduct ablation study by removing or replacing a subset of our full model. Details of different baselines are described as follows:
	
	\emph{Without self-supervised joint attention learning} ~(\textbf{Without SA}). To illustrate the contribution of the self-supervised joint attention learning in our model, we remove this part and re-train the remaining model.
	
	
	
	
	\emph{Self-supervised attention learning with CNN-based features}. To examine the contribution of attention information from either first- or third-person video in joint attention learning, we replace channel attention vectors of first-person video and third-person video with original CNN-based feature maps respectively, denoted as \textbf{CNN SA /1} and \textbf{CNN SA /3}. Then the attention is learned by minimizing the L2 distance between average pooling of the feature maps and channel attention vector of the corresponding third-person video (or first-person video).
	
	\emph{Triplet loss with CNN-based features}. To examine the contribution of attention information from either first- or third-person video in shared representation learning, we use CNN-based features of first-person video or third-person video to calculate triplet loss respectively. These two baselines denote as \textbf{CNN TL /1} and \textbf{CNN TL /3}.

	\emph{Triplet weight based on low-level image features} (\textbf{Low-level TW}) We use low-level image feature (the average gradient of image in our experiment) instead of high-level CNN-based features to estimate triplet weight.

	\begin{figure*}
		\begin{center}
			\includegraphics[width=1\linewidth]{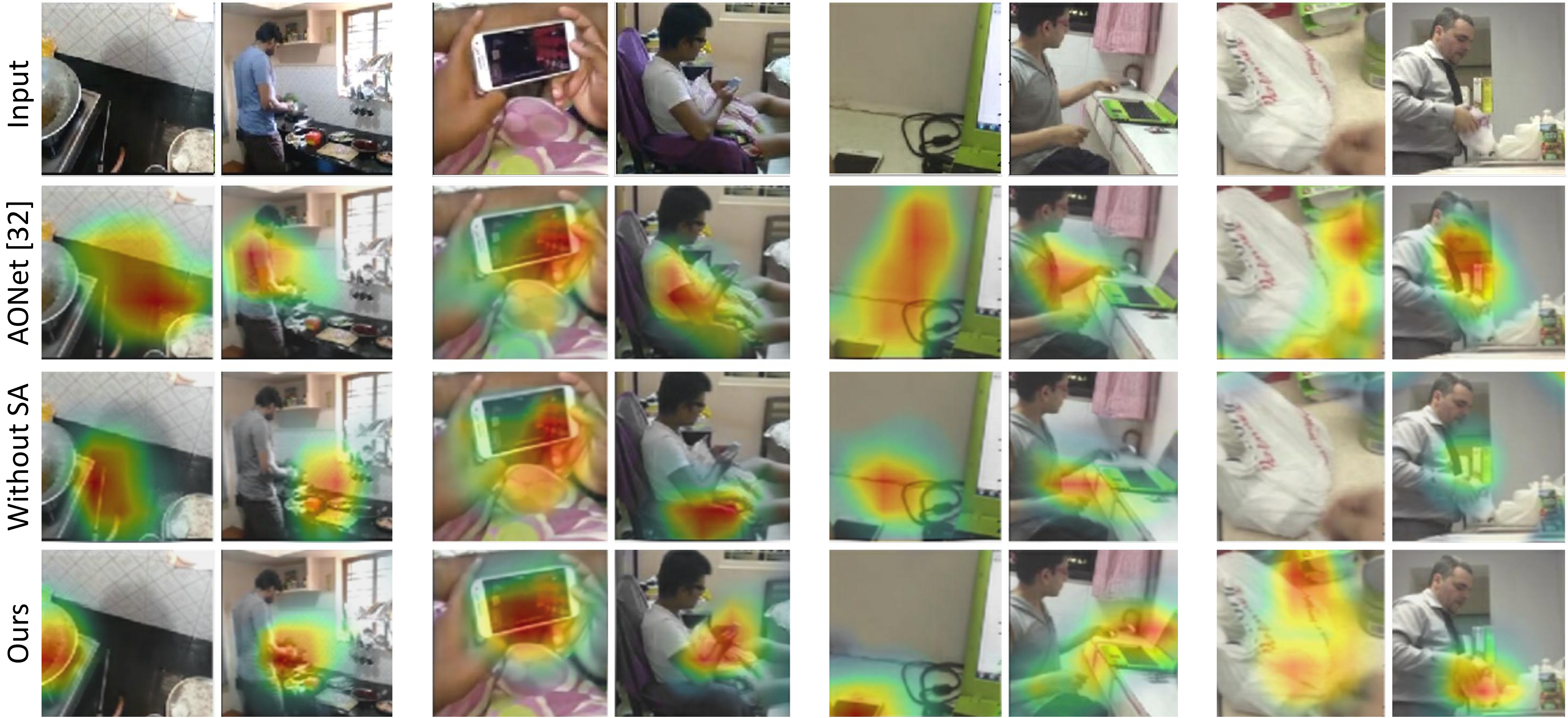}
		\end{center}
		\caption{Visualization of the predicted ROAs from AONet~\cite{sigurdsson2018actor}, \textbf{Without SA} baseline and our full model. We demonstrate four groups of image pairs, each containing frames from the first- and third-person viewpoints respectively. ROAs are visualized with heatmaps on input images. The color ranges from blue to red, showing low to high importance.}
		\label{QualitativeBaselines}
	\end{figure*}

	\begin{table}
		\setlength{\abovecaptionskip}{-0.1cm}   
		\setlength{\belowcaptionskip}{0.3cm}   
		\begin{center}
			\begin{tabular}{lcc}
				\toprule
				Method & Classification accuracy & Alignment error   \\
				\midrule
				AONet~\cite{sigurdsson2018actor}  &  51.8 & 6.5\\
				\midrule
				Without SA &  52.1 & 6.8\\
				CNN SA /1 &  88.4 & 5.2\\
				CNN SA /3 &  61.1 & 5.0\\
				CNN TL /1 & 52.4 & 6.3 \\
				CNN TL /3 & 56.9 &  6.2\\
				Low-level TW & 85.7 & 7.1\\
				\textbf{Our Full Model} & \textbf{90.6} & \textbf{4.5} \\
				\bottomrule
			\end{tabular}
		\end{center}
		\caption{Quantitative results of two tasks. The performance of pairs discrimination task is measured by classification accuracy (in \%). The performance of best-match moment localization task is measured by alignment error (in seconds) between best-match moment and ground-truth moment.}
		\label{QuantitativeResults}
		
		\vspace{-0.5cm}  
	\end{table}

	\subsection{Quantitative Analysis}
	
	
	The quantitative results of different methods on the two tasks of pairs discrimination and best-match moment localization are given in Table~\ref{QuantitativeResults}. 
	It can be seen that our method significantly outperforms AONet~\cite{sigurdsson2018actor} on both two tasks. The performance improvement probably owes to the predicted joint attention regions in shared representation learning. While AONet~\cite{sigurdsson2018actor} attempted to learn the shared representation directly from CNN-based features, our method learns the shared representation in a more efficient and reliable way by exploiting the joint attention regions which capture the same action recorded from two viewpoints.
	
	The ablation study results are also shown in the lower part of Table~\ref{QuantitativeResults}. It can be seen that the removal of attention information from either side of first- or third-person video in triplet loss leads to obvious performance drop (close to \cite{sigurdsson2018actor}), demonstrating the effectiveness of attention information in shared representation learning. Moreover, the removal of attention information from third-person video in self-supervised attention learning leads to more performance drop than similar removal from first-person video, which indicates that the attention information of third-person video plays a more important role. Most importantly, the performance degrades dramatically for \textbf{Without SA} when attention is learned independently from both viewpoints, indicating the critical role of the self-supervised joint attention learning in our full model. Overall, the ablation study results show that in shared representation learning not only the attention information is needed but also the attention in different viewpoints should be learned jointly.

	\subsection{Qualitative Analysis}
	
	Qualitative results are shown in Fig.~\ref{QualitativeBaselines}. We visualize the ROAs predicted by AONet~\cite{sigurdsson2018actor}, one of our baselines (Without SA), and our full model. We visualize the last convolutional layer activations of AONet to show which regions the network focuses on. It can be seen that~\cite{sigurdsson2018actor} tends to focus on either the image center or the visually salient object. Taking the first group (column 1 - 2 of Fig.~\ref{QualitativeBaselines}) for example, while~\cite{sigurdsson2018actor} focuses on the center region of the first-person image and human body of the third-person image, our method successfully locates the shared ROAs in both images around the object of a saucepan.
	
	As for the baseline of \textbf{Without SA} which learns attention independently for first- and third-person videos, its learned attention becomes unreliable and fails to predict the ROAs shared between two viewpoints. Taking the second group (column 3 - 4 of Fig.~\ref{QualitativeBaselines}) for example, the person is playing a mobile game, and the shared ROAs are around the region of mobile phone and hands. However, the ROA of the third-person image predicted by \textbf{Without SA} is located on the person's leg unrelated to the performed action. 
	Overall, these results demonstrate the joint attention learning via self-supervised is essential for shared representation learning between first- and third-person videos.

	\subsection{Extended Applications} 
	
	
	The goal of our method is to effectively link the first- and third-person videos. The key idea is the proposed ``joint attention" which matches the two viewpoints. Our proposed method successfully achieves this goal and its effectiveness has been proven in the previous sections.
	In addition, being able to predict and making use of the shared ROAs in the scene, our method also provides novel solutions to tackle a slice of traditional tasks.
	
	In particular, in this section we conduct experiments on three extended applications of our method, namely gaze prediction, video summarization and co-segmentation. For each application, we compare our method with the state-of-the-art quantitatively or qualitatively. Note that in the case of gaze prediction and co-segmentation, our method needs no additional input compared with traditional single view-based methods; while for video summarization, our method takes into account additional information from the first-person viewpoint and proves the significant benefit to use the proposed joint attention.

	\begin{figure}
		\setlength{\belowcaptionskip}{-0.5cm}   
		\begin{center}
			\includegraphics[width=1.0\linewidth]{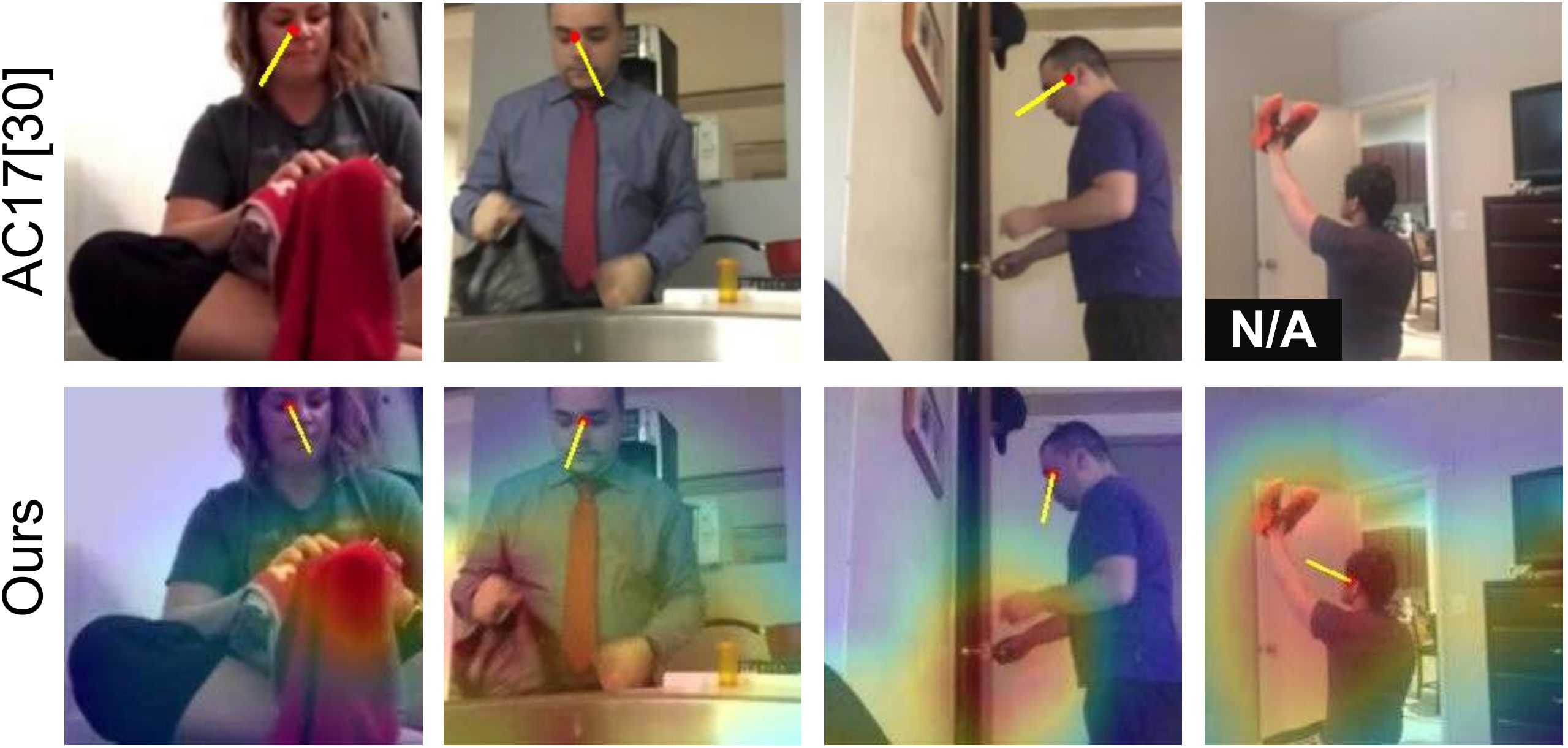}
		\end{center}
		\caption{Examples of gaze prediction results. The top row indicates the gaze prediction results of AC17~\cite{recasens2017following}. The bottom row indicates the gaze prediction results of our approach. N/A denotes AC17~\cite{recasens2017following} fails to predict the direction of human gaze. }
		\label{extendGaze}
		
	\end{figure}
	
	\subsubsection{Gaze prediction}
	Predicting where a person looks (gaze) in a third-person video is important for video analysis and information retrieval. Existing single-view based gaze prediction methods~\cite{recasens2017following,Cheng2018Appearance,Cao2017Realtime} locate the human eyes first, and then estimate the gaze direction based on the pupil and head's direction.
	Different from previous methods, we predict the gaze as follows: Firstly, we locate the position of human head based on the pose estimation method ~\cite{Cao2017Realtime}. Next, we use the third-person frame's ROA, which is inferred from our model, as the place where a person looks. 
	
	We compare our method with the state-of-the-art single-view based method AC17~\cite{recasens2017following} with its default parameters. As illustrated in Fig.~\ref{extendGaze}, without explicit gaze annotations for training, our method achieves even better results than the state-of-the-art gaze prediction method. More interestingly, our method could robustly predict a gaze position even when the human face cannot be detected in the input image. More results are provided in the supplementary.

	\begin{figure*}
		\begin{center}
			\includegraphics[width=1.0\linewidth]{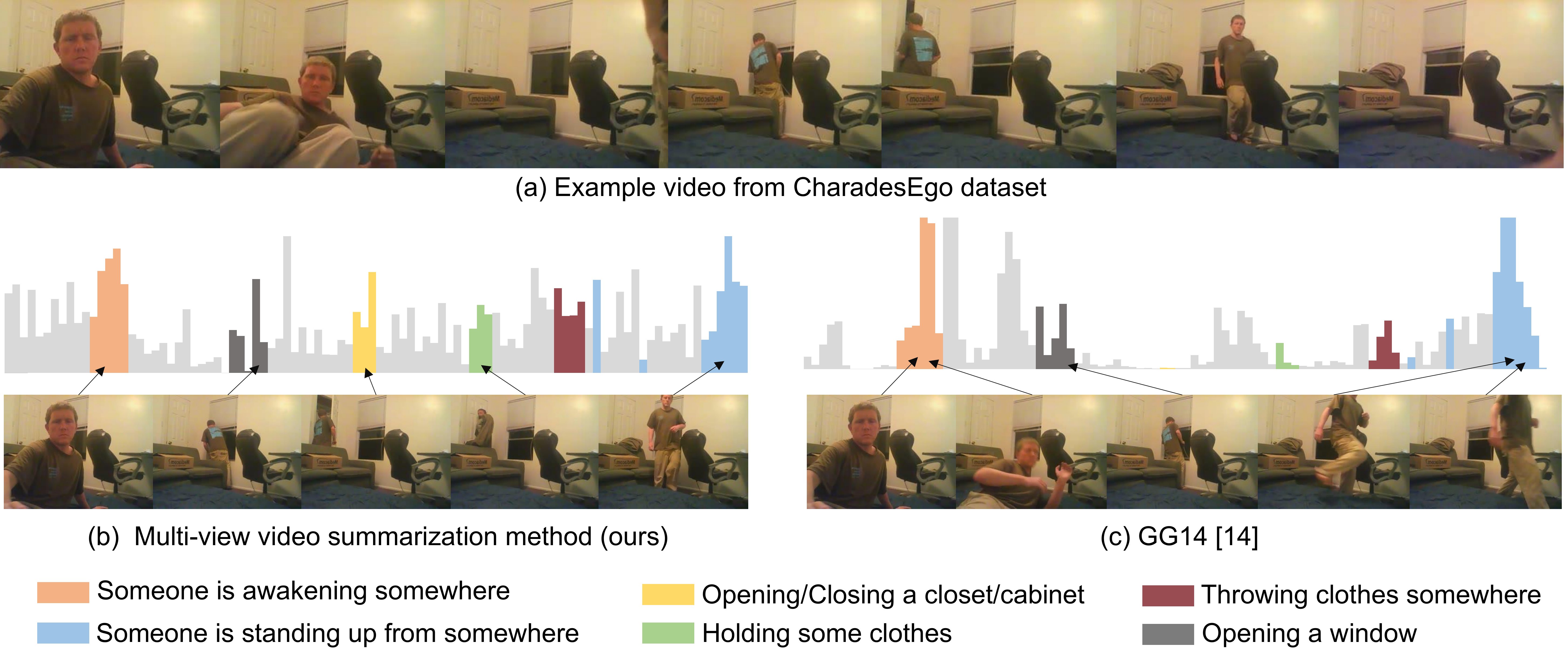}
		\end{center}
		\caption{Examples of video summaries generated by our method (b) and GG14~\cite{Gygli2014Creating} (c) on Charades-Ego dataset. Different colors of highlight bars indicate frames of different activities, while the gray colored areas correspond to the important scores. The details of action labels are shown on the bottom.}
		\label{VideoSummary}
	\end{figure*}
	
	\begin{table}
		\setlength{\abovecaptionskip}{-0.1cm}   
		\setlength{\belowcaptionskip}{0.3cm}   
		\begin{center}
			\begin{tabular}{lccc}
				\toprule
				Method & Recall  & precision  & F-scores    \\
				
				\midrule
				ZQ18~\cite{zhou2018reinforcevsumm} & 45.0 & 65.3 & 53.2\\
				GG14~\cite{Gygli2014Creating} & 56.4 & 52.9 & 54.6 \\
				\textbf{Ours} & \textbf{70.4} & \textbf{67.4} & \textbf{68.4} \\
				\bottomrule
			\end{tabular}
		\end{center}
		\caption{Quantitative comparison with state-of-the-art video summarization methods on Charades-Ego dataset. We use recall, precision and F-scores (in \%) as evaluation metrics following ~\cite{Ke2016Video}. }
		\label{VideoSumi}
		
		\vspace{-0.5cm}  
	\end{table}
	
	\subsubsection{Video summarization} In this part, we show that joint attention predicted from first- and third-person videos could be exploited to discover important moments that potentially describe the undergoing activity and thus provides an effective way for multi-view video summarization. The state-of-the-art video summarization methods~\cite{zhou2018reinforcevsumm,Vasudevan2017Query,Jin2017ElasticPlay,Zhao2017Hierarchical,Gygli2014Creating} often take single-view video as input, estimate importance scores per frame and create a video summary consisting of a small subset of frames. 
	In contrast, we evaluate per-frame confidence scores of joint attention for paired first- and third-person videos by computing the similarity between channel attention vectors of two videos. We assign high importance scores to frames that have high confidence scores of joint attention, based on which a subset of frames with high importance scores above a threshold are selected as a summary. 
	
	We compare our multi-view video summarization method with ZQ18~\cite{zhou2018reinforcevsumm} and GG14~\cite{Gygli2014Creating} on Charades-Ego dataset. We use the annotated action clips as ground-truth of video summary and use F1 score as the evaluation metric. Quantitative results are shown in Table~\ref{VideoSumi}.   
	We show qualitative comparison in Fig.~\ref{VideoSummary}. Since GG14~\cite{Gygli2014Creating} (c) predicts frame importance scores based on specific image features such as facial landmark detectors and motion, it fails to detect the important frames when the man turns his back to the camera. In contrast, our method (b) could reliably detect more important moments by joint attention information with the help from additional first-person viewpoint.

	\begin{figure}
		\begin{center}
			\includegraphics[width=1.0\linewidth]{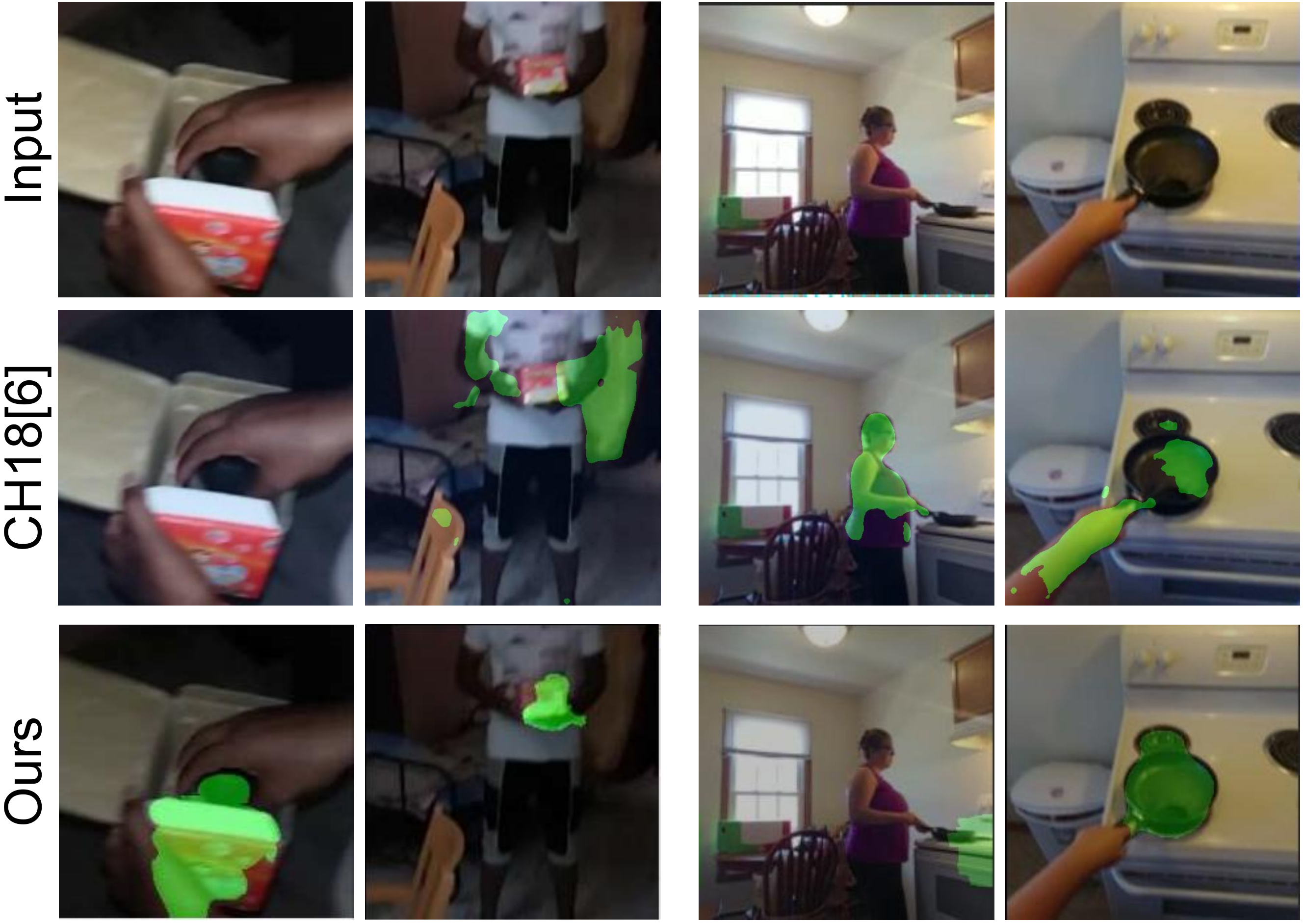}
		\end{center}
		\caption{Examples of co-segmentation results. The first row depicts input image pairs from two perspectives, the second row displays the object co-segmentation results of CH18~\cite{Hong2018Semantic}, and the third row displays the results of our approach.}
		\label{Cosegmentation}
		\vspace{-0.55cm}  
		\setlength{\belowcaptionskip}{-0.5cm}   
	\end{figure}

	\subsubsection{Co-segmentation} 
	
	Here we show that our method could be easily extended to solve co-segmentation task. 
	Different from previous methods~\cite{Hong2018Semantic,Rong2016Object,Prerana2018Object},
	the key of our approach is that we utilize the predicted ROAs of both first- and third-person videos to guide co-segmentation.
	Firstly, candidate segments could be extracted by any unsupervised image segmentation method, and we adopt an efficient superpixel algorithm SLICO~\cite{Radhakrishna2012SLI}.
	Then, we choose two segments from candidate segments of two videos as the co-segmentation outputs which are near the center of ROAs and are visually similar to each other.
	
	we compare our method with the state-of-the-art method of CH18~\cite{Hong2018Semantic}.
	As shown in Fig.~\ref{Cosegmentation}, with the guidance of attention information, our method obviously outperforms \cite{Hong2018Semantic}, even without any supervised training for co-segmentation. More results are provided in the supplementary.

	
	\section{Conclusion and future work}
	In this paper, we propose a method to effectively learn a shared representation for co-analysis of the first- and third-person videos. Our key idea is to learn joint attention to link these two viewpoints, with the assumption that shared representation should correspond to the joint attention regions. A novel representation learning framework with a self-supervised attention learning module is developed to learn joint attention between first- and third-person videos. Experiment results on a public dataset show that our proposed method significantly outperforms the state-of-the-art method on two cross-view video matching tasks. Additional experiments are conducted to demonstrate the benefits of our work for various applications. 
	
	As for our future work, we plan to explore new applications based on our current work. One possible and interesting application might be to synthesize first-person video based on the input of a third-person video.

	\bibliographystyle{plain}
	\bibliography{sample-base}

\begin{thebibliography}{10}

\bibitem{Radhakrishna2012SLI}
Radhakrishna Achanta, Appu Shaji, Kevin Smith, Aur{\'{e}}lien Lucchi, Pascal
  Fua, and Sabine S{\"{u}}sstrunk.
\newblock Slic superpixels compared to state-of-the-art superpixel methods.
\newblock {\em {IEEE} Trans. Pattern Anal. Mach. Intell.}, pages 2274--2282,
  2012.

\bibitem{Anderson2018Bottom-Up}
Peter Anderson, Xiaodong He, Chris Buehler, Damien Teney, Mark Johnson, Stephen
  Gould, and Lei Zhang.
\newblock Bottom-up and top-down attention for image captioning and visual
  question answering.
\newblock In {\em The IEEE Conference on Computer Vision and Pattern
  Recognition (CVPR)}, June 2018.

\bibitem{ardeshir2016ego2top}
Shervin Ardeshir and Ali Borji.
\newblock Ego2top: Matching viewers in egocentric and top-view videos.
\newblock In {\em European Conference on Computer Vision}, pages 253--268.
  Springer, 2016.

\bibitem{Bo2018Multi}
Zhi-Qi Cheng Hao Liu Zequn Jie Jiashi~Feng Bo~Zhao, Xiao~Wu.
\newblock Multi-view image generation from a single-view.
\newblock In {\em ACM international conference on Multimedia}, pages 383--391,
  2018.

\bibitem{Cao2017Realtime}
Zhe Cao, Tomas Simon, Shih-En Wei, and Yaser Sheikh.
\newblock Realtime multi-person 2d pose estimation using part affinity fields.
\newblock In {\em The IEEE Conference on Computer Vision and Pattern
  Recognition (CVPR)}, July 2017.

\bibitem{Hong2018Semantic}
Hong Chen, Yifei Huang, and Hideki Nakayama.
\newblock Semantic aware attention based deep object co-segmentation.
\newblock {\em ACCV}, 2018.

\bibitem{Chen2018Boosted}
Shi Chen and Qi~Zhao.
\newblock Boosted attention: Leveraging human attention for image captioning.
\newblock In {\em The European Conference on Computer Vision (ECCV)}, September
  2018.

\bibitem{Chen2018Factual}
Tianlang Chen, Zhongping Zhang, Quanzeng You, Chen Fang, Zhaowen Wang, Hailin
  Jin, and Jiebo Luo.
\newblock ``factual'' or ``emotional'': Stylized image captioning with adaptive
  learning and attention.
\newblock In {\em The European Conference on Computer Vision (ECCV)}, September
  2018.

\bibitem{Cheng2018Appearance}
Yihua Cheng, Feng Lu, and Xucong Zhang.
\newblock Appearance-based gaze estimation via evaluation-guided asymmetric
  regression.
\newblock In {\em The European Conference on Computer Vision (ECCV)}, September
  2018.

\bibitem{Chu2017Multi}
Xiao Chu, Wei Yang, Wanli Ouyang, Cheng Ma, Alan~L. Yuille, and Xiaogang Wang.
\newblock Multi-context attention for human pose estimation.
\newblock In {\em The IEEE Conference on Computer Vision and Pattern
  Recognition (CVPR)}, July 2017.

\bibitem{Du2018Interaction}
Yang Du, Chunfeng Yuan, Bing Li, Lili Zhao, Yangxi Li, and Weiming Hu.
\newblock Interaction-aware spatio-temporal pyramid attention networks for
  action classification.
\newblock In {\em The European Conference on Computer Vision (ECCV)}, September
  2018.

\bibitem{fan2017identifying}
Chenyou Fan, Jangwon Lee, Mingze Xu, Krishna Kumar~Singh, Yong Jae~Lee, David~J
  Crandall, and Michael~S Ryoo.
\newblock Identifying first-person camera wearers in third-person videos.
\newblock In {\em Proceedings of the IEEE Conference on Computer Vision and
  Pattern Recognition}, pages 5125--5133, 2017.

\bibitem{Jun2019Dual}
Jun Fu, Jing Liu, Haijie Tian, Zhiwei Fang, and Hanqing Lu.
\newblock Dual attention network for scene segmentation.
\newblock {\em CVPR}, abs/1809.02983, 2019.

\bibitem{Gygli2014Creating}
Michael Gygli, Helmut Grabner, Hayko Riemenschneider, and Luc Van~Gool.
\newblock Creating summaries from user videos.
\newblock In {\em ECCV}, 2014.

\bibitem{Kaiming2016Deep}
Kaiming He, Xiangyu Zhang, Shaoqing Ren, and Jian Sun.
\newblock Deep residual learning for image recognition.
\newblock {\em CVPR}, 2016.

\bibitem{Hu2018Squeeze}
Jie Hu, Li~Shen, and Gang Sun.
\newblock Squeeze-and-excitation networks.
\newblock In {\em The IEEE Conference on Computer Vision and Pattern
  Recognition (CVPR)}, June 2018.

\bibitem{Hu2018CVM}
Sixing Hu, Mengdan Feng, Rang M.~H. Nguyen, and Gim Hee~Lee.
\newblock Cvm-net: Cross-view matching network for image-based ground-to-aerial
  geo-localization.
\newblock In {\em The IEEE Conference on Computer Vision and Pattern
  Recognition (CVPR)}, June 2018.

\bibitem{itti1998model}
Laurent Itti, Christof Koch, and Ernst Niebur.
\newblock A model of saliency-based visual attention for rapid scene analysis.
\newblock {\em IEEE Transactions on Pattern Analysis \&amp; Machine
  Intelligence}, (11):1254--1259, 1998.

\bibitem{Jin2017ElasticPlay}
Haojian Jin, Yale Song, and Koji Yatani.
\newblock Elasticplay: Interactive video summarization with dynamic time
  budgets.
\newblock In {\em Proceedings of the 25th ACM International Conference on
  Multimedia}, MM '17, 2017.

\bibitem{Lejbolle2018Attention}
Aske~R. Lejbolle, Benjamin Krogh, Kamal Nasrollahi, and Thomas~B. Moeslund.
\newblock Attention in multimodal neural networks for person re-identification.
\newblock In {\em The IEEE Conference on Computer Vision and Pattern
  Recognition (CVPR) Workshops}, June 2018.

\bibitem{Li2018Diversity}
Shuang Li, Slawomir Bak, Peter Carr, and Xiaogang Wang.
\newblock Diversity regularized spatiotemporal attention for video-based person
  re-identification.
\newblock In {\em The IEEE Conference on Computer Vision and Pattern
  Recognition (CVPR)}, June 2018.

\bibitem{Li2018Harmonious}
Wei Li, Xiatian Zhu, and Shaogang Gong.
\newblock Harmonious attention network for person re-identification.
\newblock In {\em The IEEE Conference on Computer Vision and Pattern
  Recognition (CVPR)}, June 2018.

\bibitem{Y2018Multi}
Y.~{Li} and Y.~{Wang}.
\newblock A multi-label image classification algorithm based on attention
  model.
\newblock In {\em 2018 IEEE/ACIS 17th International Conference on Computer and
  Information Science (ICIS)}, pages 728--731, June 2018.

\bibitem{Wentao2018Cascaded}
Wentao Liu, Jie Chen, Cheng Li, Chen Qian, Xiao Chu, and Xiaolin Hu.
\newblock A cascaded inception of inception network with attention modulated
  feature fusion for human pose estimation, 2018.

\bibitem{Prerana2018Object}
Prerana Mukherjee, Brejesh Lall, and Snehith Lattupally.
\newblock Object cosegmentation using deep siamese network.
\newblock 2018.

\bibitem{Parisotto2018Global}
Emilio Parisotto, Devendra Singh~Chaplot, Jian Zhang, and Ruslan Salakhutdinov.
\newblock Global pose estimation with an attention-based recurrent network.
\newblock In {\em The IEEE Conference on Computer Vision and Pattern
  Recognition (CVPR) Workshops}, June 2018.

\bibitem{paszke2017automatic}
Adam Paszke, Sam Gross, Soumith Chintala, Gregory Chanan, Edward Yang, Zachary
  DeVito, Zeming Lin, Alban Desmaison, Luca Antiga, and Adam Lerer.
\newblock Automatic differentiation in pytorch.
\newblock In {\em NIPS-W}, 2017.

\bibitem{Y2018Object}
Y.~{Peng}, X.~{He}, and J.~{Zhao}.
\newblock Object-part attention model for fine-grained image classification.
\newblock {\em IEEE Transactions on Image Processing}, March 2018.

\bibitem{Rong2016Object}
Rong Quan, Junwei Han, Dingwen Zhang, and Feiping Nie.
\newblock Object co-segmentation via graph optimized-flexible manifold ranking.
\newblock In {\em 2016 {IEEE} Conference on Computer Vision and Pattern
  Recognition, {CVPR} 2016, Las Vegas, NV, USA, June 27-30, 2016}, pages
  687--695, 2016.

\bibitem{recasens2017following}
Adria Recasens, Carl Vondrick, Aditya Khosla, and Antonio Torralba.
\newblock Following gaze in video.
\newblock In {\em Proceedings of the IEEE International Conference on Computer
  Vision}, pages 1435--1443, 2017.

\bibitem{Regmi2018Cross}
Krishna Regmi and Ali Borji.
\newblock Cross-view image synthesis using conditional gans.
\newblock In {\em The IEEE Conference on Computer Vision and Pattern
  Recognition (CVPR)}, June 2018.

\bibitem{sigurdsson2018actor}
Gunnar~A Sigurdsson, Abhinav Gupta, Cordelia Schmid, Ali Farhadi, and Karteek
  Alahari.
\newblock Actor and observer: Joint modeling of first and third-person videos.
\newblock In {\em Proceedings of the IEEE Conference on Computer Vision and
  Pattern Recognition}, pages 7396--7404, 2018.

\bibitem{Gunnar2016Hollywood}
Gunnar~A. Sigurdsson, G{\"{u}}l Varol, Xiaolong Wang, Ali Farhadi, Ivan Laptev,
  and Abhinav Gupta.
\newblock Hollywood in homes: Crowdsourcing data collection for activity
  understanding.
\newblock {\em ECCV}, abs/1604.01753, 2016.

\bibitem{Song2018Mask}
Chunfeng Song, Yan Huang, Wanli Ouyang, and Liang Wang.
\newblock Mask-guided contrastive attention model for person re-identification.
\newblock In {\em The IEEE Conference on Computer Vision and Pattern
  Recognition (CVPR)}, June 2018.

\bibitem{Vasudevan2017Query}
Arun~Balajee Vasudevan, Michael Gygli, Anna Volokitin, and Luc Van~Gool.
\newblock Query-adaptive video summarization via quality-aware relevance
  estimation.
\newblock In {\em Proceedings of the 25th ACM International Conference on
  Multimedia}, pages 582--590, 2017.

\bibitem{Nam2016Localizing}
James Vo, Nam N.and~Hays.
\newblock Localizing and orienting street views using overhead imagery.
\newblock In {\em Computer Vision -- ECCV 2016}, pages 494--509, Cham, 2016.
  Springer International Publishing.

\bibitem{Wang2017Residual}
Fei Wang, Mengqing Jiang, Chen Qian, Shuo Yang, Cheng Li, Honggang Zhang,
  Xiaogang Wang, and Xiaoou Tang.
\newblock Residual attention network for image classification.
\newblock In {\em The IEEE Conference on Computer Vision and Pattern
  Recognition (CVPR)}, July 2017.

\bibitem{Wang2018Non-Local}
Xiaolong Wang, Ross Girshick, Abhinav Gupta, and Kaiming He.
\newblock Non-local neural networks.
\newblock In {\em The IEEE Conference on Computer Vision and Pattern
  Recognition (CVPR)}, June 2018.

\bibitem{Woo2018CBAM}
Sanghyun Woo, Jongchan Park, Joon-Young Lee, and In~So~Kweon.
\newblock Cbam: Convolutional block attention module.
\newblock In {\em The European Conference on Computer Vision (ECCV)}, September
  2018.

\bibitem{xu2018joint}
Mingze Xu, Chenyou Fan, Yuchen Wang, Michael~S Ryoo, and David~J Crandall.
\newblock Joint person segmentation and identification in synchronized
  first-and third-person videos.
\newblock In {\em Proceedings of the European Conference on Computer Vision
  (ECCV)}, pages 637--652, 2018.

\bibitem{yonetani2015ego}
Ryo Yonetani, Kris~M Kitani, and Yoichi Sato.
\newblock Ego-surfing first-person videos.
\newblock In {\em Proceedings of the IEEE Conference on Computer Vision and
  Pattern Recognition}, pages 5445--5454, 2015.

\bibitem{Zhang2018deep}
Jing Zhang, Tong Zhang, Yuchao Dai, Mehrtash Harandi, and Richard Hartley.
\newblock Deep unsupervised saliency detection: A multiple noisy labeling
  perspective.
\newblock In {\em The IEEE Conference on Computer Vision and Pattern
  Recognition (CVPR)}, June 2018.

\bibitem{Ke2016Video}
Ke~Zhang, Wei{-}Lun Chao, Fei Sha, and Kristen Grauman.
\newblock Video summarization with long short-term memory.
\newblock In {\em Computer Vision - {ECCV} 2016 - 14th European Conference,
  Amsterdam, The Netherlands, October 11-14, 2016, Proceedings, Part {VII}},
  pages 766--782, 2016.

\bibitem{Zhao2017Hierarchical}
Bin Zhao, Xuelong Li, and Xiaoqiang Lu.
\newblock Hierarchical recurrent neural network for video summarization.
\newblock In {\em Proceedings of the 25th ACM International Conference on
  Multimedia}, MM '17, pages 863--871, 2017.

\bibitem{zhou2018reinforcevsumm}
Kaiyang Zhou, Yu~Qiao, and Tao Xiang.
\newblock Deep reinforcement learning for unsupervised video summarization with
  diversity-representativeness reward.
\newblock {\em AAAI}, 2018.

\end{thebibliography}
	
	\begin{figure*}
		\begin{center}
			\Huge \textbf{What I See Is What You See: Joint Attention Learning for First and Third Person Video Co-analysis\\\emph{(supplementary material)}}\\
		\end{center}
	\end{figure*}

	\newpage

	\appendix
	\section{Supplementary Overview}
	
	This supplementary material presents additional results related to extended experiments. We first provide more qualitative results of Section 5.4.1 and Section 5.4.3. 
	Then we evaluate the perceptual quality of our generated images by conducting an human subjective evaluation.
	More details are present in the following sections.
	

	\section{More results of extended experiments}
	\subsection{Gaze Prediction}
\begin{figure*}
		\begin{center}
			\includegraphics[width=\linewidth,height = 21.2cm]{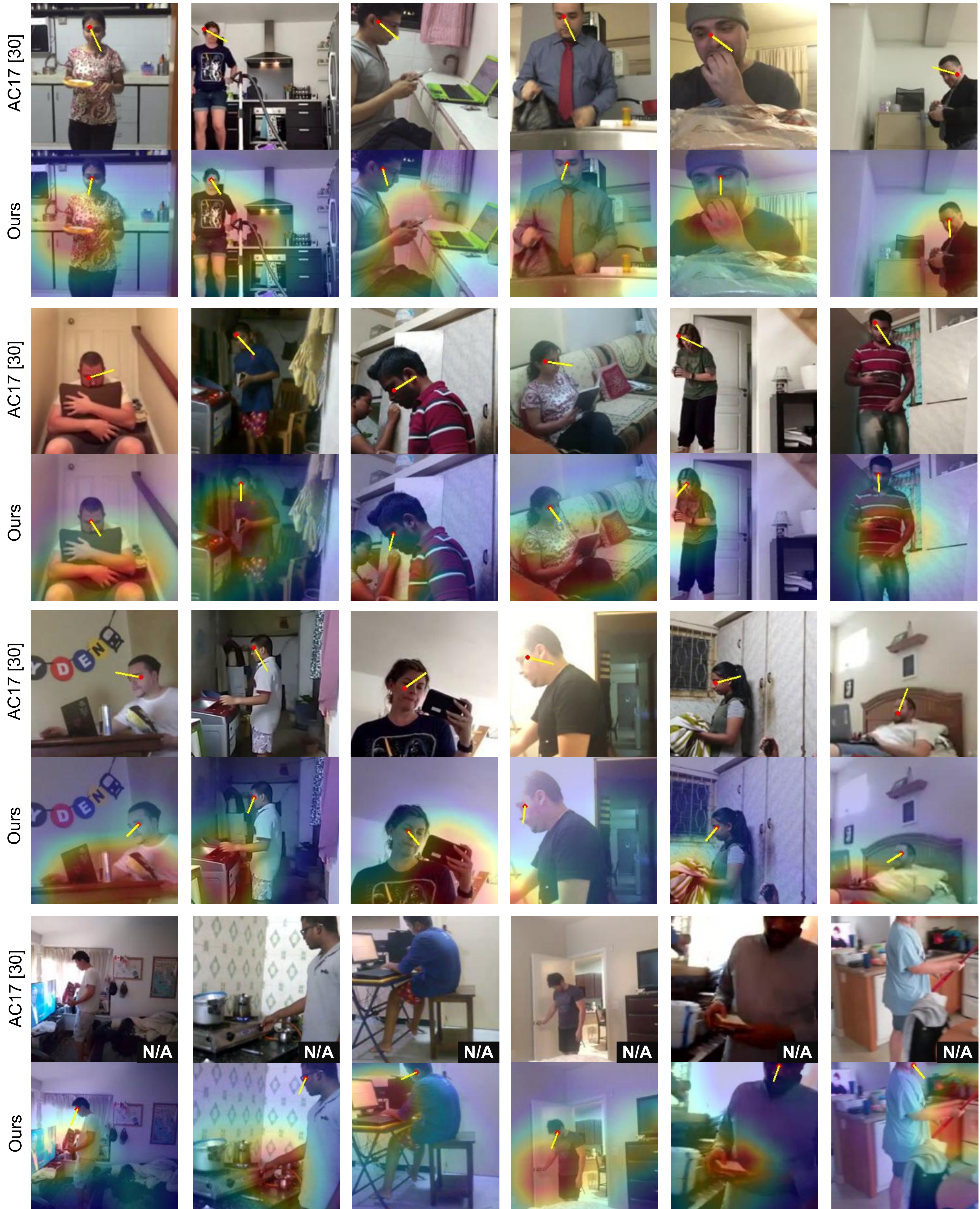}
		\end{center}
		\caption{Gaze prediction results compare with AC17~\cite{recasens2017following}. N/A denotes AC17~\cite{recasens2017following} fails to predict the direction of human gaze. }
		\label{gaze1} 
	\end{figure*}

	Following up the analysis of Section 5.4.1 in the paper, the third-person frame's ROA is used to predict gaze direction in videos. Our method achieves even better results than the state-of-the-art gaze prediction method, and even can predict a robust gaze position when the human face cannot be detected. We show in Fig. \ref{gaze1} some examples demonstrating such improvement on gaze direction. 
	\subsection{Co-segmentation}
\begin{figure*}
		\begin{center}
			\includegraphics[width=\linewidth]{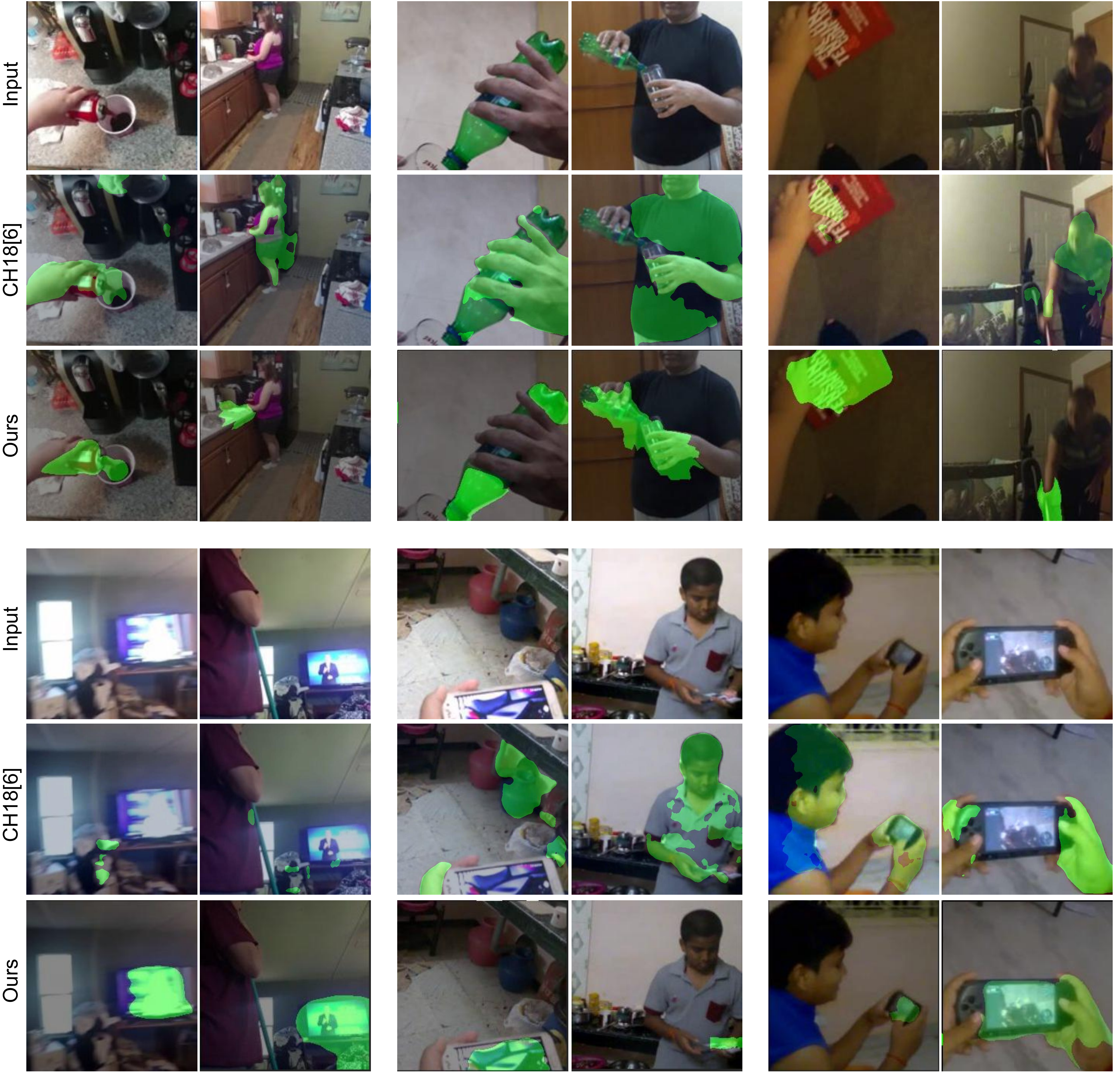}
		\end{center}
		\caption{Co-segmentation results compared with CH18~\cite{Hong2018Semantic}.}
		\label{co-seg} 
		
	\end{figure*}

	Since the first-person view region only corresponds to a small and deformed part of that in the third-person view, it's difficult for state-of-the-art method to predict segment region. We use the predicted ROAs of both first- and third-person videos to guide co-segmentation, compared with state-of-the-art method of CH18 \cite{Hong2018Semantic}. More results are shown in Fig. \ref{co-seg}. 
	
	\begin{table}[b]
		\setlength{\abovecaptionskip}{0.3cm}   
		\setlength{\belowcaptionskip}{0.3cm}   
		\begin{center}
			\begin{tabular}{lc}
				\toprule
				& Preference rate \\
				\midrule
				Ours > AC17~\cite{recasens2017following} & 92.3 \\
				Ours > CH18~\cite{Hong2018Semantic} &  84.6 \\
				\bottomrule
			\end{tabular}
		\end{center}
		\caption{The results of subjective evaluation (in \%). AC17~\cite{recasens2017following} is the state-of-the-art method of gaze prediction. CH18~\cite{Hong2018Semantic} is the state-of-the-art method of co-segmentation. The preference rate shows that our results are strongly preferred over the state-of-the-art methods}
		\label{userstudy}
	\end{table}

	\section{Subjective evaluation}
	This section provides the details of our subjective evaluation on gaze prediction application and co-segmentation application respectively. We conduct blind randomized A/B testing against gaze prediction method AC17~\cite{recasens2017following} and co-segmentation method CH18~\cite{Hong2018Semantic}. In total, 13 workers participated in the experiment. We show participants 24 samples in each task. All the test frames are from Charades-Ego dataset. The results are listed in Table \ref{userstudy}.
	
	\subsection{Human Subjective Evaluation on Gaze Prediction}
	We present three images in a row: an input third-person frame, our gaze prediction result (denoted as $\textbf{A}$), and the gaze prediction result from AC17~\cite{recasens2017following} we compared with (denoted as $\textbf{B}$). We ask the question: ``$\textbf{A}$ and $\textbf{B}$ are two version of gaze prediction result of the given input. The red dot presents the eyes location and the yellow line presents the gaze direction. Which result has better performance?''

	\subsection{Human Subjective Evaluation on Co-segmentation}
	
	We present three pair of first- and third-person frames in column to conduct human subjective evaluation on object co-segmentation. The first row contain original input pair of first-person frame and its corresponding third-person frame, the second row demonstrate our co-segmentation result (denoted as $\textbf{A}$), and the third row is the result of CH18 \cite{Hong2018Semantic} (denoted as $\textbf{B}$). We ask the question: ``$\textbf{A}$ and $\textbf{B}$ are two version of co-segmentation result of the given input. The co-segment regions are highlight in green. Which result has better performance?''

\end{document}